\documentclass[letterpaper, 10 pt, conference]{ieeeconf}  

\IEEEoverridecommandlockouts                              

\usepackage{amsmath} 
\usepackage{amssymb}  

\usepackage{epsfig} 
\usepackage{hyperref}
\usepackage{multirow}
\usepackage{booktabs}
\usepackage{threeparttable}
\usepackage{bm}
\usepackage{multicol}
\usepackage[table,xcdraw]{xcolor}
\usepackage{tikz}
\usepackage{algorithmic}
\usepackage{algorithm}
\usepackage[caption=false,font=small,labelfont=rm,textfont=rm]{subfig}
\usepackage{adjustbox}
\usepackage{colortbl}
\usepackage{xcolor}  
\usepackage{cite}

\title{\LARGE \bf
A Data-Driven Aggressive Autonomous Racing Framework Utilizing Local Trajectory Planning with Velocity Prediction
}
\author{Zhouheng Li$^{1}$, Bei Zhou$^{1}$, Cheng Hu$^{1}$, Lei Xie$^{1, \dagger}$, Hongye Su$^{1}$
\thanks{This work was supported by the Ningbo Key research and development Plan (No.2023Z116).}
\thanks{$\dagger$ Corresponding author: {\tt\small leix@iipc.zju.edu.cn}.}
\thanks{$^{1}$ State Key Laboratory of Industrial, Zhejiang University, Hangzhou 310027, China.
        {\tt\small \{zh.li,zhoubei,22032081\}@zju.edu.cn;
        	\{leix,hysu\}@iipc.zju.edu.cn}.}%
}
\begin{document}

\maketitle
\thispagestyle{empty}
\pagestyle{empty}

\begin{abstract}

The development of autonomous driving has boosted the research on autonomous racing. However, existing local trajectory planning methods have difficulty planning trajectories with optimal velocity profiles at racetracks with sharp corners, thus weakening the performance of autonomous racing. To address this problem, we propose a local trajectory planning method that integrates Velocity Prediction based on Model Predictive Contouring Control (VPMPCC). The optimal parameters of VPMPCC are learned through Bayesian Optimization (BO) based on a proposed novel Objective Function adapted to Racing (OFR). Specifically, VPMPCC achieves velocity prediction by encoding the racetrack as a reference velocity profile and incorporating it into the optimization problem. This method optimizes the velocity profile of local trajectories, especially at corners with significant curvature. The proposed OFR balances racing performance with vehicle safety, ensuring safe and efficient BO training. In the simulation, the number of training iterations for OFR-based BO is reduced by 42.86\% compared to the state-of-the-art method. The optimal simulation-trained parameters are then applied to a real-world F1TENTH vehicle without retraining. During prolonged racing on a custom-built racetrack featuring significant sharp corners, the mean projected velocity of VPMPCC reaches 93.18\% of the vehicle's handling limits. The released code is available at \textcolor{blue}{https://github.com/zhouhengli/VPMPCC}.

\end{abstract}

\section{INTRODUCTION}

\begin{figure}[!t]
	\centering
	\includegraphics[scale=0.3]{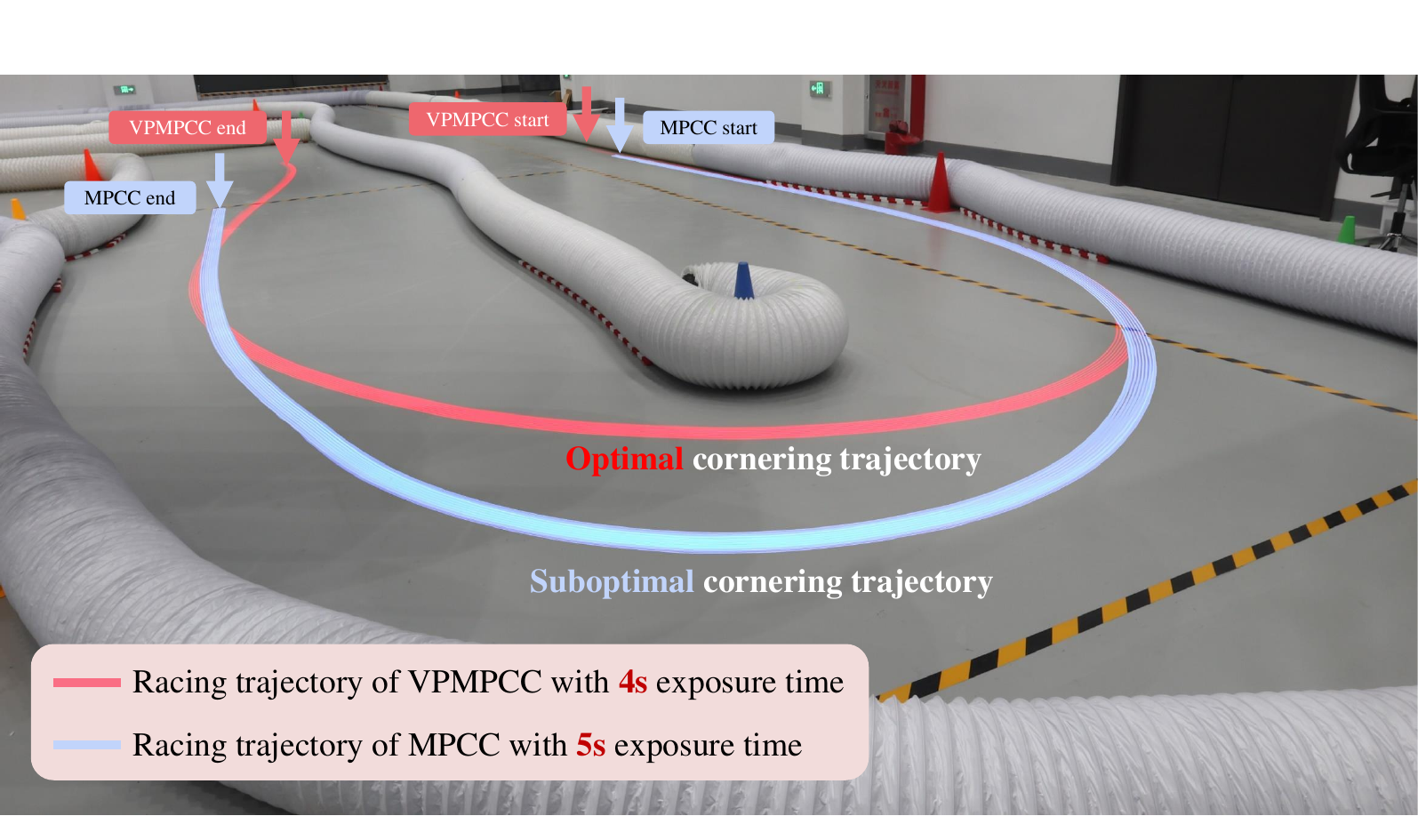}
	\caption{Comparison of trajectories between the proposed Velocity Prediction MPCC (VPMPCC) and standard MPCC through a high-curvature corner using a self-built F1TENTH vehicle. The trajectories are captured through long exposures. VPMPCC is able to plan optimal cornering trajectories at higher velocities, resulting in shorter cornering time to reduce lap time.}
	\label{fig:teaser}
\end{figure}

Autonomous racing stems from the development of autonomous driving. Pushing vehicle performance to its limits is the goal of autonomous racing. This poses a significant challenge to local trajectory planning methods due to the following four factors: (a) minimum lap time guarantee \cite{christTimeoptimalTrajectoryPlanning2021,huNovelModelPredictive2024,huCombinedFastControl2022}
; (b) control feasibility of the planned trajectory\cite{xuAutonomousVehicleMotion2021,fanBaiduApolloEM2018,liRapidIterativeTrajectory}; (c) computational efficiency of the optimization problem\cite{schulmanMotionPlanningSequential2014,hanEfficientSpatialTemporalTrajectory2024,mao2024overtaking}; and (d) safe and efficient parameters tuning\cite{frohlichContextualTuningModel2022,krinner2024time,zhou2025adaptive}. Global trajectory planning\cite{christTimeoptimalTrajectoryPlanning2021,heilmeier2020minimum} leverages complete racetrack information for racing trajectory planning. However, it fails to account for real-time vehicle state changes, causing control issues and thus reducing racing performance. Recently, local trajectory planning methods integrating planning and control have been widely studied\cite{lyonsCurvatureAwareModelPredictive2023,linigerOptimizationBasedAutonomous2015,kabzanAMZDriverlessFull2020,kabzan2019learning,li2025reduce}.  Model Predictive Contouring Control (MPCC) is a typical example of such an approach\cite{linigerOptimizationBasedAutonomous2015,kabzanAMZDriverlessFull2020}. To realize minimum lap time, MPCC introduces the projected velocity as an effective metric to represent the racing progress. This method performs well on racetracks without significant curvature corners\cite{kabzanAMZDriverlessFull2020}. However, with sharp corners on the racetrack (as shown in Fig.~\ref{fig:teaser}), relying solely on tracking error to balance the  projected velocity maximization makes it difficult for MPCC to plan cornering trajectories with optimal velocities\cite{betzAutonomousVehiclesEdge2022}.  To tackle this problem, we propose a Velocity Prediction MPCC (VPMPCC) local trajectory planning method. Specifically, the proposed VPMPCC realizes velocity prediction by using vehicle longitudinal velocity as an independent velocity decision variable. The key advantage lies in optimizing the velocity profile of the local trajectory, particularly on corners with significant curvature, ultimately reducing lap time.

In recent years, data-driven Bayesian optimization (BO)\cite{frohlichContextualTuningModel2022,krinner2024time} has been developed to realize automatic parameters tuning. However, these methods do not consider modeling vehicle safety while reducing lap time, resulting in the inability to guarantee vehicle safety during generalization\cite{berkenkampBayesianOptimizationSafety2023}. Meanwhile, simultaneously ensuring the accuracy of the vehicle model and achieving computational efficiency of the optimization problem is also a primary challenge\cite{fanBaiduApolloEM2018,schulmanMotionPlanningSequential2014}.  To bridge these gaps, a safe and efficient data-driven racing framework is proposed, incorporating the VPMPCC and BO. Specifically, in closed-loop BO training, the proposed framework uses the high-fidelity vehicle dynamics model obtained from offline system identification to filter the racing trajectory planned by the vehicle kinematic model-based VPMPCC. Moreover, a practical Objective Function adapted to Racing (OFR) is designed for BO to achieve efficient training. The advantage of this framework is that the parameters of the VPMPCC planner can be safely trained in the simulation and transferred to real-world vehicles without retraining. Therefore, the vehicle performance can be pushed to its limits efficiently and safely. In summary, this paper provides the following contributions:

\begin{enumerate}
	
	\item \textbf{An effective method for integrating velocity prediction to efficiently optimize local trajectory velocity profiles.} The proposed VPMPCC integrates velocity prediction into trajectory planning by encoding the racetrack as a Reference Velocity Profile (RVP). Velocity prediction is then achieved by matching the planned velocities within the prediction horizon with the corresponding reference velocities in the RVP.
		
	\item \textbf{An effective OFR accelerates BO to safely learn the optimal parameters of the local trajectory planner.} The proposed OFR enhances vehicle safety by limiting racing trajectory deviations significantly from the reference line and reduces the number of iterations required for convergence by applying negative costs to time-reducing trajectories. It improves training efficiency by \textbf{42.86\%}, with optimal simulation parameters transferable to real-world vehicles.
	
	\item \textbf{{A data-driven autonomous racing framework safely pushes the vehicle to its limits, validated through reliable real-world experiments.}} {The VPMPCC and BO-based autonomous racing framework (VPBO-RF) outperform on a self-built 1:10 scale F1TENTH vehicle, excelling on a challenging racetrack with sharp corners.} Despite limited computational resources and execution delays, the mean projected velocity of VPMPCC achieved \textbf{93.18\% }of vehicle handling limits, with an efficient mean computation time of \textbf{7.04 ms}. 
\end{enumerate}

This paper consists of several sections. Section \ref{Preliminaries} introduces the system dynamic of VPMPCC and conventional MPCC, followed by an overview of BO. Section \ref{sec:VPMPCC} introduces the implementation of VPMPCC. Section \ref{sec:VPBO-RF} describes the overall framework of VPBO-RF, focusing on the design of the OFR. Section \ref{sec:Experimental} includes simulations and real-world experiments, while Section \ref{sec:Conclusion} serves as the conclusion, summarizing the essential findings and outlining directions for future research.

\section{Preliminaries}\label{Preliminaries}

This section presents the system dynamics of VPMPCC, an introduction to MPCC, and an overview of BO.

\subsection{System Dynamics}
In contrast to the conventional MPCC, VPMPCC decouples the projected velocity $ v_{\textrm{p}} $ from the vehicle longitudinal velocity $ v $. The advantage of this design lies in the introduction of velocity prediction, enabling a trade-off between maximizing the projected velocity, thus handling corners with significant curvature. Therefore, the system dynamics used by VPMPCC is:
\begin{equation}\label{eq:f_zeta_u}
	\dot{\bm{\zeta}}=f(\bm{\zeta},\mathbf{u}) \stackrel{\text{def}}{=} \quad
	\begin{aligned}
		\dot{x}  &= \cos (\varphi) \cdot {v}\\ 
		\dot{y}&=  \sin (\varphi) \cdot {v}\\ 
		\dot{\varphi}&=  {{v}} \cdot \tan(\delta)\cdot L^{-1}\\
		\dot{s}&={v}_{\textrm{p}} \\
	\end{aligned}
\end{equation}
where $\varphi$ is the yaw angle and $L$ is the wheelbase. Then let  $\bm{\zeta}= [x ,\, y ,\, \varphi ,\, s]^{\top} $ denote the state space, and $ \mathbf{u} = [v ,\, \delta ,\, v_{\textrm{p}}]^{\top} $ denotes the configuration space. The final control inputs for the underlying chassis are $v$ and steering angle $ \delta $. Therefore, the weight matrix $\mathbf{R}$ in \eqref{eq:cost_mpcc} is expressed as $ \mathrm{diag} \left( \begin{bmatrix} q_{\Delta v} & q_{\Delta \delta} & {q}_{\Delta v_{\textrm{p}}} \end{bmatrix} \right) $.

\subsection{Model Predictive Contouring Control}

\begin{figure}[!t]
	\centering 
	\includegraphics[scale=0.28]{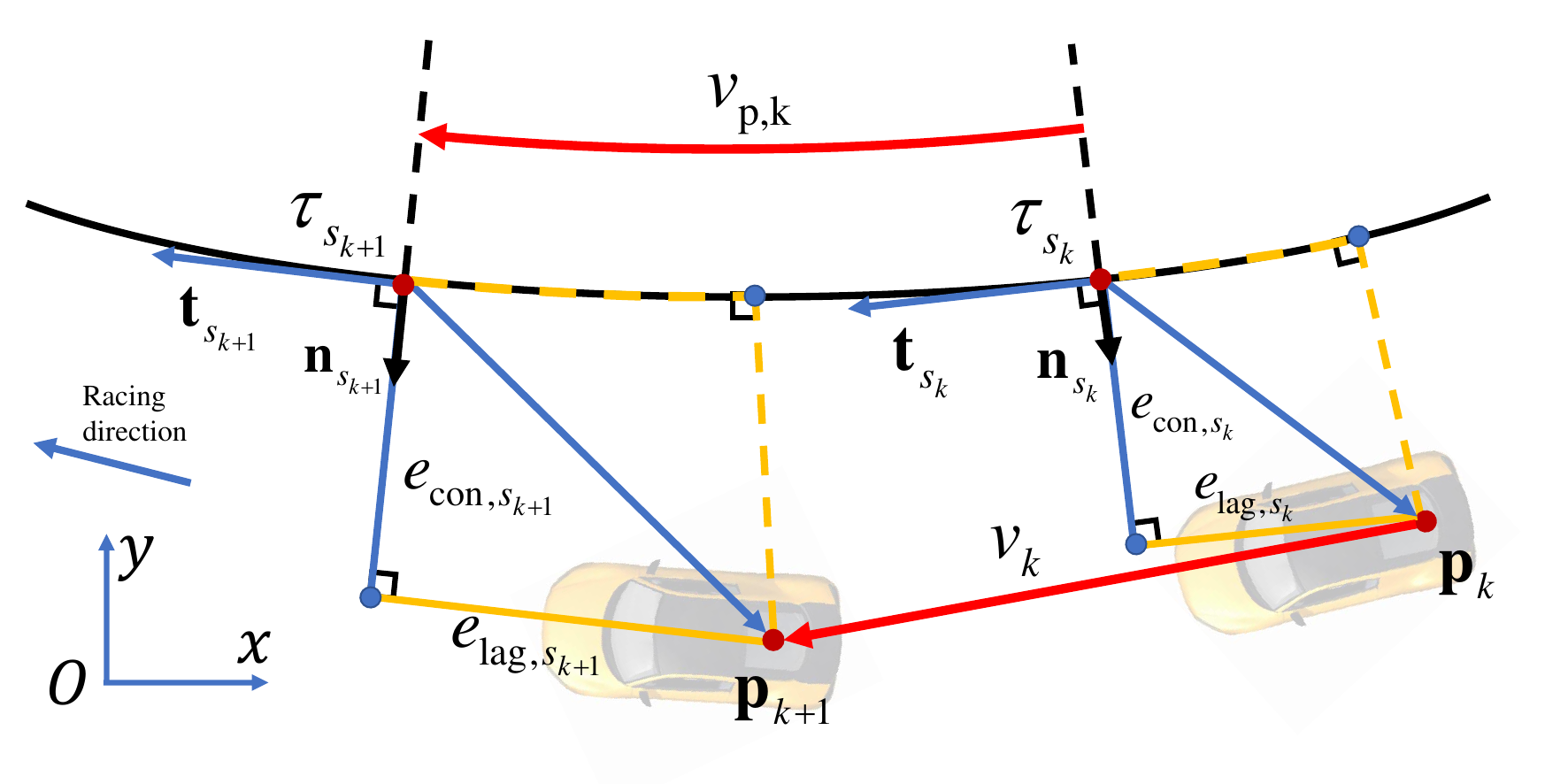}
	\caption{Schematic diagram illustrating the calculation of contouring and lag errors in conventional MPCC. The proposed VPMPCC method incorporates the vehicle's longitudinal velocity as an independent decision variable.}
	\label{fig:mpcc}
\end{figure}

\begin{figure*}[!t]
	\centering
	\includegraphics[scale=0.46]{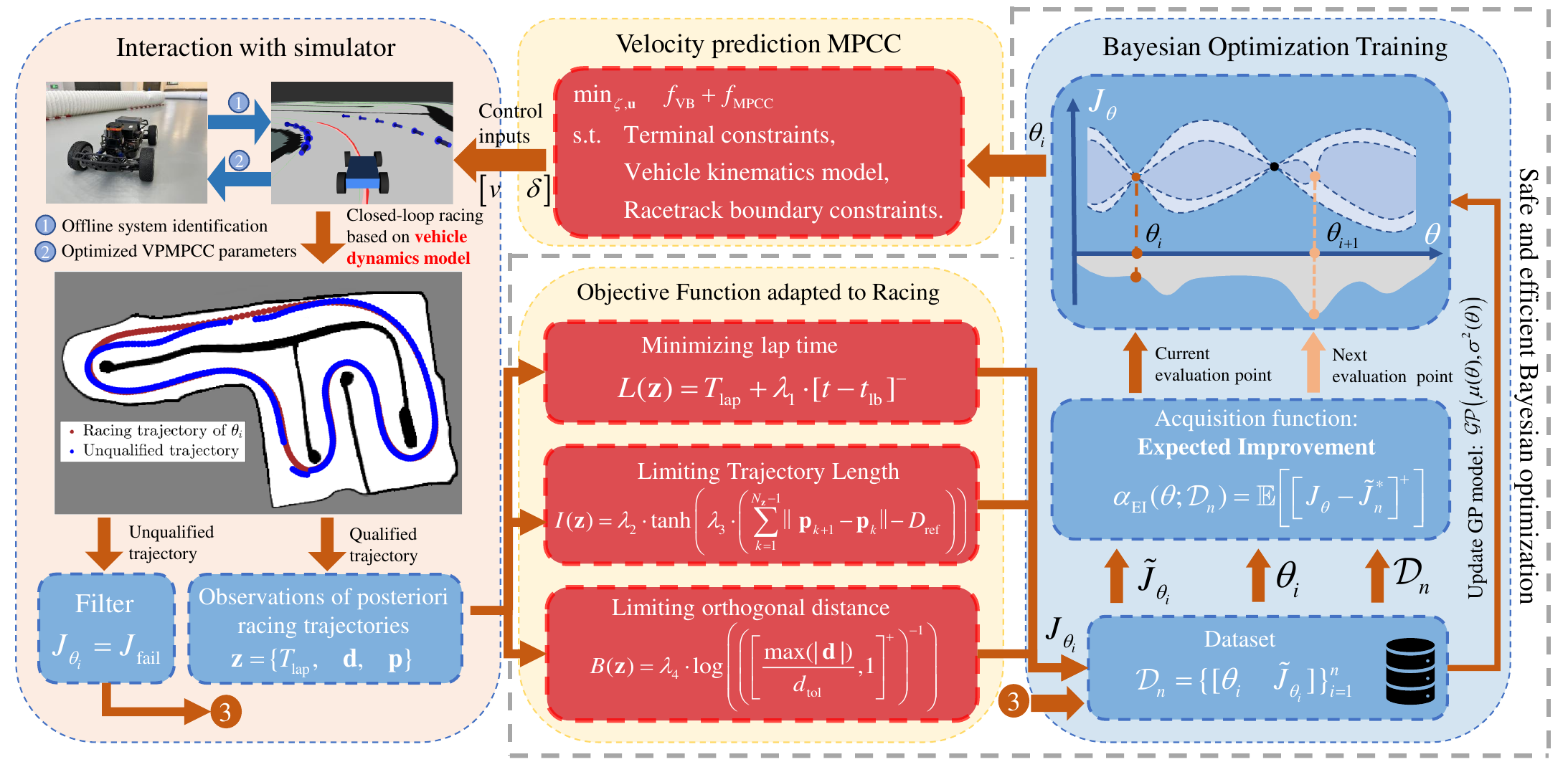}
	\caption{
		Schematic diagram of the VPBO-RF.  The \textcolor[RGB]{197,90,17}{\textbf{brown}} arrows indicate the online evaluation process of parameter $\bm{\theta}_i$. The \textcolor[RGB]{244,177,131}{\textbf{light brown}} arrows indicate the next evaluation parameter $\bm{\theta}_{i+1}$ chosen by the acquisition function. The closed-loop training convergence means that the OFR value no longer decreases.}
	\label{fig:pipline}
\end{figure*} 
As shown in Fig.~\ref{fig:mpcc}, the contouring error $e_{\textrm{con}}$ and the lag error $e_{\textrm{lag}}$ of MPCC\cite{linigerOptimizationBasedAutonomous2015,lyonsCurvatureAwareModelPredictive2023} are calculated as follows:
\begin{equation}
	\begin{aligned}\label{eq:state_w}
		&e_{\textrm{con},s} = \mathbf{n}_{s} \cdot 
		(\mathbf{p}	- \bm{\tau}_{s}),\,
		e_{\textrm{lag},s} = \mathbf{t}_{s} \cdot 
		(\mathbf{p}	- \bm{\tau}_{s})\\
		&\mathbf{e}_{s}={\begin{bmatrix} e_{\textrm{con},s} 
				& e_{\textrm{lag},s} \end{bmatrix}}^{\top},\,
		\mathbf{c} = {\begin{bmatrix} (e_{\textrm{con}}^{\textrm{max}})^{-1} & (e_{\textrm{lag}}^{\textrm{max}})^{-1} \end{bmatrix}}			
	\end{aligned}
\end{equation}
where $\mathbf{p} = \begin{bmatrix} x & y \end{bmatrix}$ represents the position of the rear axle center in the Cartesian frame. The reference line, denoted by $\bm{\tau}$, is parameterized by the arc length $s$. For a given arc length $s$, $\mathbf{n}_{s}$ and $\mathbf{t}_{s}$ are the normal and tangent unit vectors of the reference point $\bm{\tau}_s=[x_{\bm{\tau},s},\,y_{\bm{\tau},s}]$. And $\mathbf{c}$ is the normalization vector for the tracking error $\mathbf{e}_s$. The MPCC approximates the projected velocity $ v_{\textrm{p}}$  to the vehicle longitudinal velocity $ v $, expressed as $ v_{\textrm{p}} \approx   v $. Then, with the help of $\mathbf{e}_{s}$ and $v_{\textrm{p}}$,  the composition of the cost function of MPCC is shown below:
\begin{equation}\label{eq:cost_mpcc}
	f_{\textrm{MPCC}} =  \sum_{k=1}^{N_p} -\frac{\gamma \cdot v_{{\textrm{p}},k}}{v_{\textrm{max}}} \cdot T_{s}   +
	\Vert \mathbf{c} \cdot  \mathbf{e}_{s_k} \Vert^2_{\mathbf{q}_{\mathbf{e}}} + 
	\lVert \mathbf{u}_{k+1} - \mathbf{u}_k \rVert^2_{\mathbf{R}} 
\end{equation}
where $N_p$ denotes the prediction horizon and  $\gamma$ is the scalar weight of maximum projected velocity. The weight matrix $\mathbf{q}_{\mathbf{e}}$ for contouring and lag errors is $ \mathrm{diag} \left( \begin{bmatrix} q_{\mathbf{e}_{\textrm{con}}} & q_{\mathbf{e}_{\textrm{lag}}} \end{bmatrix} \right)$, and the weight matrix $\mathbf{R}$ is for the change of control variables. $v_{\textrm{max}}$ is the maximum longitudinal velocity used for normalization.

\subsection{Bayesian Optimization}

The BO aims to solve the following optimization problem given a series of observations $\mathbf{z}$:
\begin{equation}
	\begin{aligned}\label{eq:BO}
		\bm{\theta}^{*}=\arg \min_{\bm{\theta} \in \Theta} J(\bm{\theta} ; \mathbf{z})
	\end{aligned}
\end{equation}
where $J_{\bm{\theta}}$ denotes the objective function, which a surrogate model approximates. $\bm{\theta}$ is the optimization variable. And $\Theta$ denotes the feasible space of all optimization parameters. The Gaussian Process (GP) is the most widely used surrogate model denoted as $J_{\bm{\theta}} \sim \mathcal{GP}\left(\mu({\bm{\theta}}),\sigma^2(\bm{\theta})\right)$, $\mu({\bm{\theta}})$ is the mean function, and $\sigma^2(\bm{\theta})=\mathcal{K}({\bm{\theta}}_i, {\bm{\theta}}_j)$ is the kernel function. To initialize the GP model, full-space random sampling is performed within the feasible space $\Theta$ to construct the  priori dataset, which is denoted as $		\mathcal{D}_n =  \left\{
\begin{bmatrix}
	\bm{\theta}_i & \tilde{J}_{\bm{\theta}_i}
\end{bmatrix}
\right\}_{i=1}^n$, where
$n$ is the number of random samples and $\tilde{J}_{\bm{\theta}_i} = {J}_{\bm{\theta}_i} + \eta $. Here, $\eta$ is the Gaussian-distributed noise with mean 0 and covariance $\sigma_{\eta}$. During the prediction of the selected evaluation point $\bm{\theta}_{i}$, the dataset $\mathcal{D}_n$ is utilized to construct the covariance matrix $\mathbf{K}_{n}= \left[ \mathcal{K}(\bm{\theta}_i, \bm{\theta}_j) \right]_{i,j=1}^n$ and covariance vector $\mathbf{k}_n = \left[ \mathcal{K}(\bm{\theta}^*, \bm{\theta}_i) \right]_{i=1}^n$.  Then, the selected evaluation points are determined using the acquisition function~\cite{shahriariTakingHumanOut2016}.

\section{Velocity Prediction MPCC}\label{sec:VPMPCC}

{This section outlines the VPMPCC implementation, emphasizing velocity prediction and the trade-off between velocity prediction and maximizing projected velocity.}

\subsection{Design of the Velocity Prediction}

The VPMPCC is proposed to optimize the velocity profile for local trajectory planning on racetracks with sharp curvature corners. The velocity prediction is accomplished by enhancing the similarity between planned velocities in the prediction horizon and the corresponding reference velocity in RVP. Therefore, the following {cost} function for velocity prediction can be constructed:
\begin{equation}\label{eq:VBP_pre}
	\begin{aligned}
		f_{\textrm{VP}}&= \frac{q_v}{v_{\Delta,\textrm{max}}} \sum_{k=1}^{N_p} \left( v_k - {v_{\textrm{RVP},{s_k}}} \right)^2 \\
	\end{aligned}
\end{equation}
where $v_{\Delta,\textrm{max}}$ denotes the maximum error in velocity matching. {The ${v_{\textrm{RVP},{s_k}}}$ represents the parameterized RVP based on arc length, and $k$ denotes the time instance within prediction horizon.} The $q_v$ is a scalar weight. VPMPCC uses the minimum curvature path of \cite{heilmeier2020minimum} as the reference line and the corresponding velocity profile as the RVP. The VPMPCC controls the racing performance through the velocity prediction component $f_{\textrm{VP}}$ and the maximizing projected velocity component in the Eq.~\eqref{eq:cost_mpcc} together. At racetrack corners with high curvature, maximizing the projected velocity component incurs a significant cost in the velocity prediction component. This forces the optimization problem to shift from the aggressive decision of maximizing the vehicle's projected velocity to the safer decision of reducing the velocity.  It is important to note that the RVP is used only for velocity prediction, not velocity tracking. This distinction is the fundamental difference between VPMPCC and longitudinal velocity tracking controllers\cite{wengAggressiveCorneringFramework2024}.

\subsection{VPMPCC Formation}
According to~\eqref{eq:VBP_pre} and \eqref{eq:cost_mpcc}, the final VPMPCC optimization problem can be summarized as:
\begin{equation}
	\begin{aligned}\label{eq:vpmpcc}
		\min_{\bm{\zeta,\mathbf{u}} } \quad & f_{\textrm{VP}} + f_{\textrm{MPCC}} \\
		\text{s.t.} \quad 
		&\bm{\zeta}_{k+1} = \bm{\zeta}_{k} + T_s \cdot f\left(\bm{\zeta}_{k},\mathbf{u}_{k}\right),\bm{\zeta}_0=\bm{\zeta}_{\textrm{cur}},\\
		&\bm{\zeta}_{\textrm{min}} \leq  \lvert \bm{\zeta}_k \rvert \leq \bm{\zeta}_{\textrm{max}},\,\mathbf{u}_{\textrm{min}} \leq  \lvert \mathbf{u}_k \rvert \leq \mathbf{u}_{\textrm{max}}.
	\end{aligned}
\end{equation}
where $\bm{\zeta}_{\textrm{max}}$ and $\bm{\zeta}_{\textrm{min}}$ denote the upper and lower bounds of the state constraints, while $\mathbf{u}_{\textrm{max}}$ and $\mathbf{u}_{\textrm{min}}$ are the upper and lower limits of the control variable, respectively. $T_s$ is the prediction interval, and $\bm{\zeta}_{\textrm{cur}}$ is current state values. The racetrack width is scaled using $\xi < 1$, ensuring the planned trajectory remains fully within the racetrack for safety. 

\section{Data-driven Autonomous Racing Framework Utilizing High-efficiency OFR-driven BO}\label{sec:VPBO-RF}
This section outlines the OFR design, focusing on racing performance and vehicle safety modeling, followed by the method for filtering abnormal trajectories using the vehicle dynamics model. The overall training framework is shown in Fig.~\ref{fig:pipline}. The vehicle dynamics model and the system identification method are detailed in~\cite{beckerModelAccelerationbasedPursuit2023}. The parameters of the proposed VPMPCC to be optimized are: 
\begin{equation*}
	\begin{aligned}\label{eq:vpmpcc_pars}
		\bm{\theta}={\begin{bmatrix} 
				N_p & q_v & \gamma & e_{\textrm{con}} & e_{\textrm{lag}}  & q_{\Delta v} & q_{\Delta \delta} & 
				{q}_{\Delta v_{\textrm{p}}} & \xi
		\end{bmatrix}}^{\top}_{9 \times 1}
	\end{aligned}
\end{equation*}
where the meaning of the parameters in $\bm{\theta} $  is described in detail in Section~\ref{Preliminaries} and ~\ref{sec:VPMPCC}.  The observations chosen for BO is  $\mathbf{z}=\{ T_{\textrm{lap}}, \,  \mathbf{d},\, \mathbf{P} \}$. $T_{\textrm{lap}}$ denotes lap time. $\mathbf{d}=[d_1,\,\cdots,\,d_{N_{\mathbf{z}}}]$ is the orthogonal distance of the racing trajectory from the reference line. $\mathbf{P}=[\mathbf{p}_1,\,\cdots,\,\mathbf{p}_{N_{\mathbf{z}}}]$ is the list of the racing trajectory points. $N_{\mathbf{z}}$ is the number of trajectory points. The number of training episodes is $N_{\textrm{BO}}$.

\begin{figure*}[htbp]
	\centering
	\subfloat[Hardware of the self-built F1TENTH vehicle]
	{
		\includegraphics[scale=0.175]{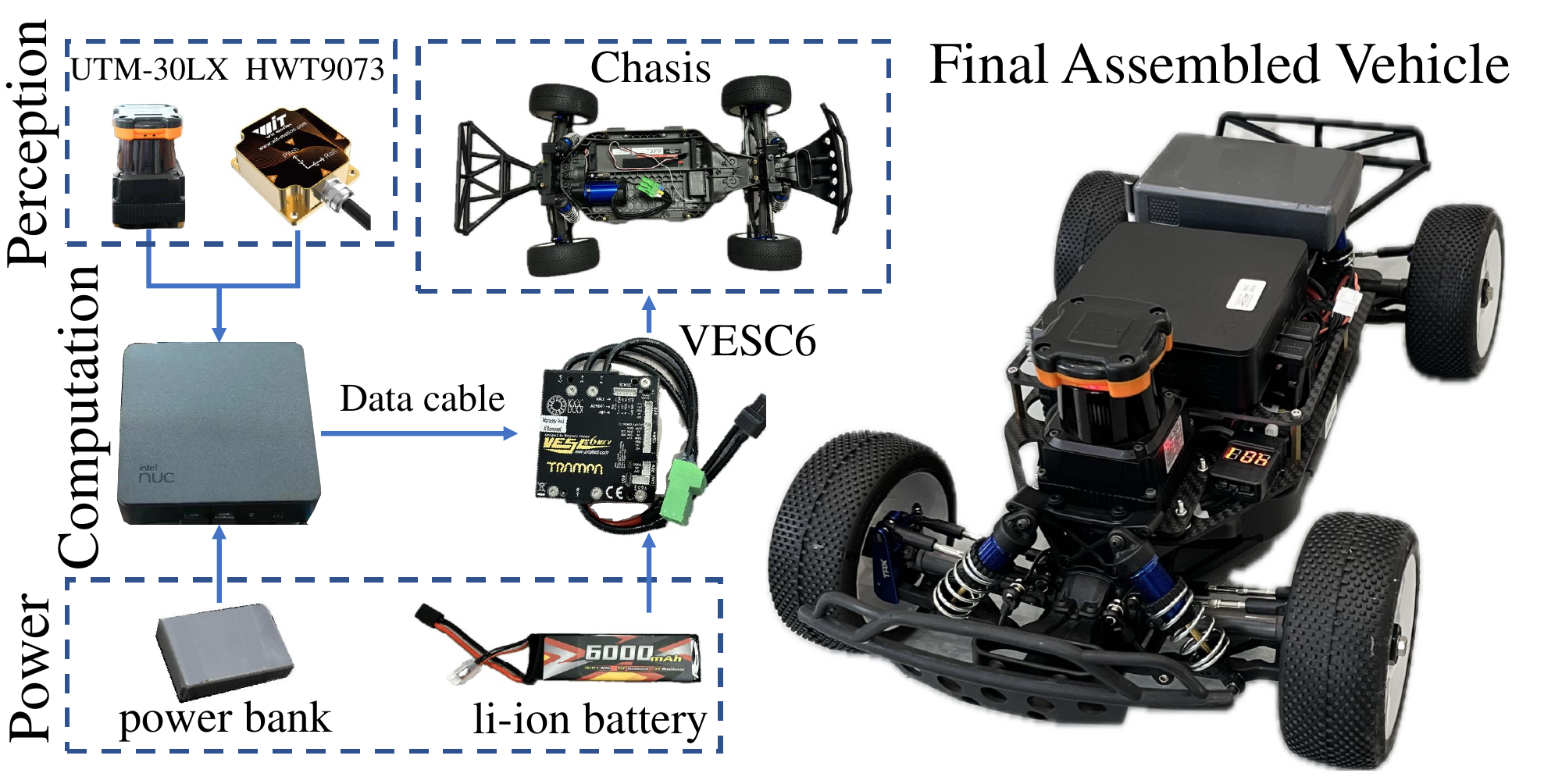}
		\label{fig:car_comp}
	}
	\vspace{0.01cm} %
	\subfloat[Racetrack includes four sharp corners]
	{
		\includegraphics[scale=0.175]{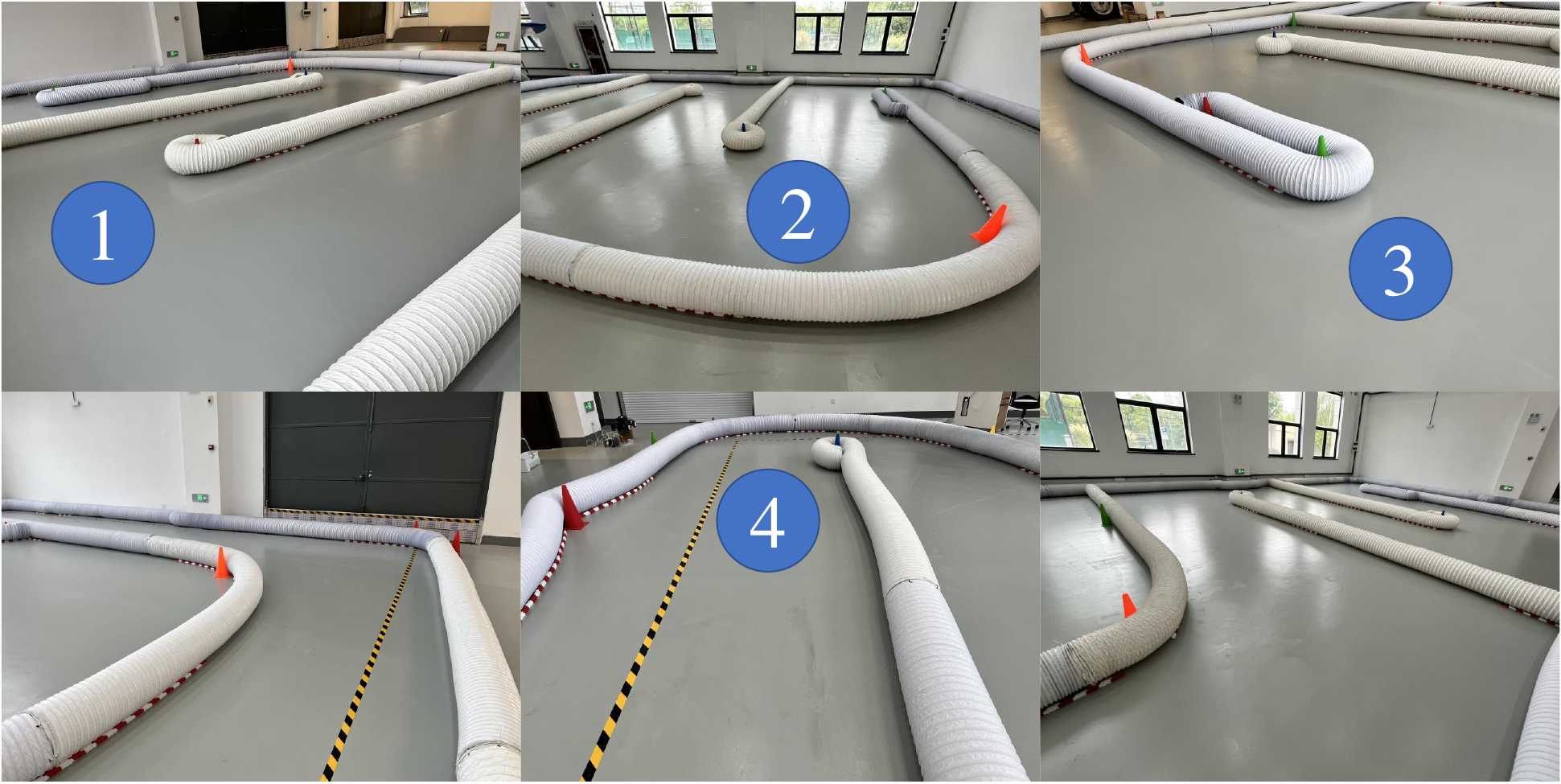}
		\label{fig:racetrack}
	}
	\vspace{0.01cm} %
	\subfloat[Grid map of the racetrack]
	{
		\includegraphics[scale=0.165]{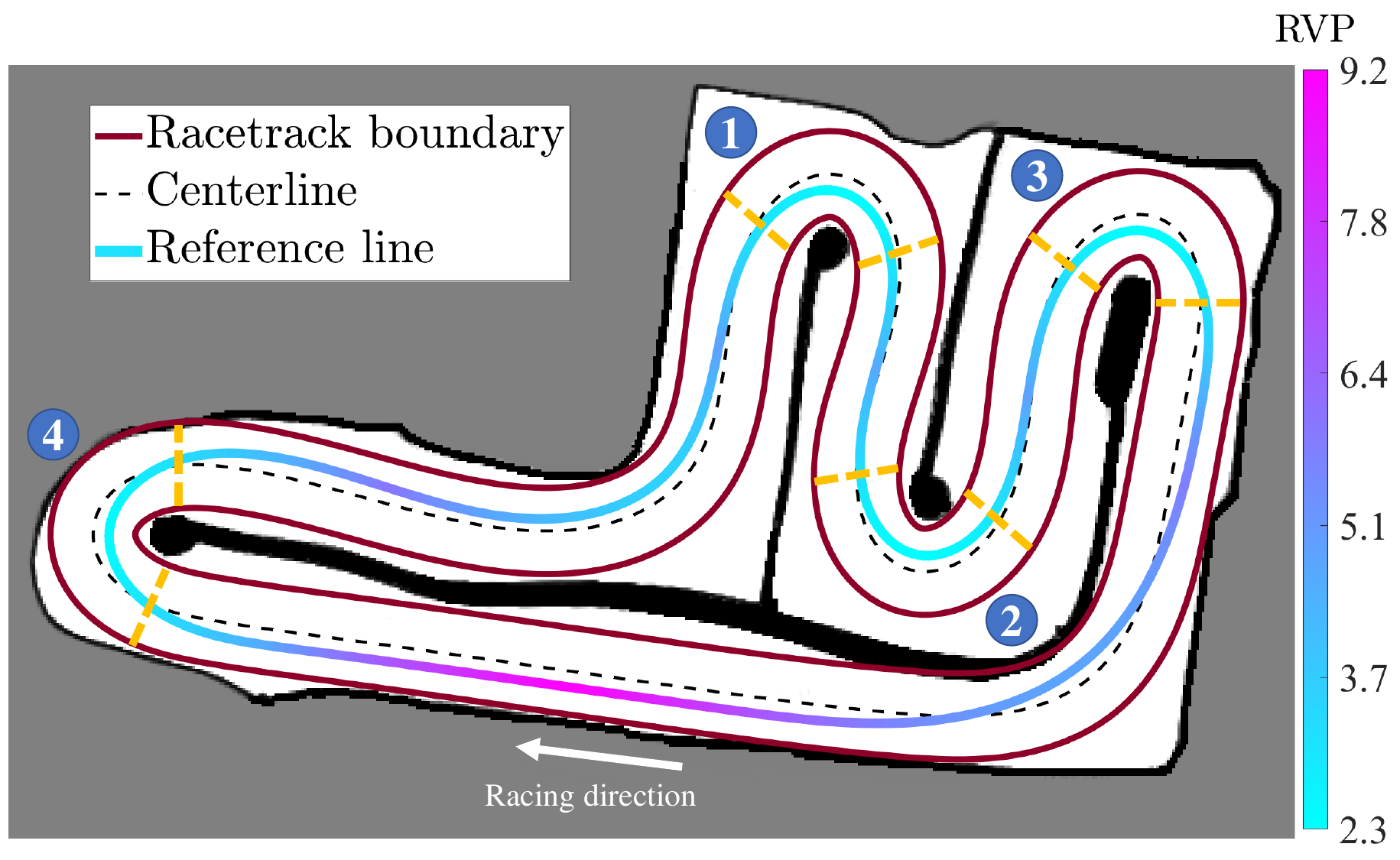}
		\label{fig:map}
	}
	\caption{The grid map created from real racetracks determines boundaries for VPMPCC and is used for localization during real vehicle racing.}
	\label{fig:real_racingtraj_car}
\end{figure*}

\subsubsection{Surrogate model}
The OFR denoted as  $J_{\bm{\theta}}$ consists of the following components: lap time, trajectory length minimization, and limitations on orthogonal distance. Firstly, the modeling of ${L(\mathbf{z})}$ encouraging VPMPCC to explore lower lap time is:
\begin{equation}
	\begin{aligned}\label{eq:laptime}
		{L(\mathbf{z})} = T_{\textrm{lap}} 
		+ 
		\lambda_1 \cdot [t-t_{\textrm{lb}}]^{-}
	\end{aligned}
\end{equation}
where $[a]^{-}=\min(a,0)$, and $t_{\textrm{lb}}$ denotes the threshold for reducing objective function value.

Then, the trajectory length minimization  is modeled as:
\begin{equation}
	\begin{aligned}\label{eq:drifts_OD}
		{I( \mathbf{z})} = \lambda_2 \cdot \tanh\left( \lambda_3 \cdot
			\left(
					\sum_{i=1}^{N_\textbf{z}-1} 
					\Vert 
					\mathbf{p}_{i+1} - \mathbf{p}_{i}
					\Vert - D_{\textrm{ref}}
			\right)
		\right)
	\end{aligned}
\end{equation}
where $\lambda_2, \lambda_3$ achieve a trade-off between different components. $ D_{\textrm{ref}}$ is the length of the reference line. The purpose of this component is to force VPMPCC to search for shorter racing trajectories, thereby accelerating the exploration of minimal lap time. 

The limitation on the orthogonal distance  of racing trajectory  using the $\log$ barrier function is presented below:
\begin{equation}
	\begin{aligned}\label{eq:OD}
		{B(\mathbf{z} )} = 
		\lambda_4 \cdot 
		\log \left(
		\left( { 
			\left[  
			\frac{ 
				\max(\lvert \mathbf{d} \rvert  ) 
			}{
				d_{\textrm{tol}}
			}  
			,1 
			\right]^{+}}  
		\right)^{-1}\right)
	\end{aligned}
\end{equation}
where $[a,1]^{+}=\max(a,1)$, $\lambda_4 < 0$, and the $ d_{\textrm{tol}}$ denotes the maximum orthogonal distance at which the vehicle is allowed to deviate from the reference line.  Normally, ${B\left(\mathbf{z}\right)}$ does not work, but it gives a large penalty when the vehicle deviates significantly. So the value of $\lvert \lambda_4 \rvert$ needs to be large.

The vehicle kinematics model cannot fully capture the dynamics of the vehicle, leading to abnormal trajectories during racing, as shown in Fig.~\ref{fig:pipline}.  This is because control inputs calculated based on the vehicle kinematics model may lead to abnormal position shifts after being integrated over time by the vehicle dynamics model. During BO training, these abnormal trajectories are identified as unqualified and penalized through ${J}_{\textrm{fail}}$, preventing further exploration of these trajectories. Therefore, the $J_{\bm{\theta}}$ can be expressed as:
\begin{equation}
	\begin{aligned}\label{eq:J}
		J_{\bm{\theta}} = 
		\begin{cases}
			J_{\textrm{fail}}, & \text{if } \Delta \mathbf{p}  \geq d_{\textrm{ub}} \text{ or } D_{\textrm{traj}} < D_{\textrm{lb}}, \\
			L(\mathbf{z}) + I(\mathbf{z}) + B(\mathbf{z}), & \text{otherwise}.
		\end{cases}
	\end{aligned}
\end{equation}
where $\Delta \mathbf{p} = \Vert \mathbf{p}_i - \mathbf{p}_{i+1} \Vert,\,\forall i \in[1,...,N_{\mathbf{z}-1}]$. The $d_{\textrm{ub}}$ denotes the maximum distance between neighboring points on the racing trajectory, and $D_{\textrm{lb}}$ denotes the minimum racing trajectory length. {These metrics are designed to penalize trajectories with position drifts.} The value of OFR for crashed racing trajectories is all ${J}_{\textrm{fail}}$.

\subsubsection{Acquisition function}

The acquisition function chosen is {Expected Improvement (EI)}, which is represented as:
\begin{equation}
	\begin{aligned}\label{eq:EI}
		\alpha_{\textrm{EI}}(\bm{\theta} ; \mathcal{D}_n) = 
		\mathbb{E}\left[
		\left[ {J}_{\bm{\theta}} - \tilde{J}_n^{*} \right]^{+}  
		\right]
	\end{aligned}
\end{equation}
where $[a]^{+}=\max{(a,0)}$, and $\tilde{J}_n^{*}$ is best value to date. {The Mat\`{e}rn kernel with parameter \( \nu = \frac{5}{2} \) is used as the kernel function\cite{zhou2025adaptive}.} The next evaluation point $\bm{\theta}_{i+1}$ is then selected using $ \bm{\theta}_{i+1} = \arg \max_{\bm{\theta} \in \Theta} \alpha_{\textrm{EI}}(\bm{\theta} ; \mathcal{D}_n)$. VPMPCC  is then reconstructed based on $\bm{\theta}_{i+1}$ for collecting new observations $\mathbf{z}_{i+1}$, which is used to enrich $\mathcal{D}_n$, thus completing the closed-loop training.

\section{EXPERIMENTAL RESULTS AND DISCUSSION}\label{sec:Experimental}

\begin{table}[b]
	\centering
	\caption{Main parameters of the proposed VPMPCC and BO-based autonomous racing framework.}
	\begin{tabular}{p{5em}p{5.2em}|p{4em}p{6.835em}}
		\toprule
		\textbf{Parameters} & \textbf{Setting} & \textbf{Parameters} & \textbf{Setting} \\
		\midrule
		\midrule
		$[N_{\textrm{BO}},v_{\Delta,\textrm{max}}]$ & {[200, 10m/s] } & $\lambda$ &  [20,10,0.5,-100]  \\
		$[L,T_s]$    & [0.32m,0.1s] & $\mathbf{u}_{\textrm{min}}$ & \multicolumn{1}{c}{[-15m/s -0.4rad -15m/s]}  \\
		$[d_{\textrm{tol}},t_{\textrm{lb}]}$ & [0.5m,17.6s] & $\mathbf{u}_{\textrm{max}}$  & \multicolumn{1}{c}{[15m/s$\;\,$ 0.4rad$\;\,$ 15m/s]} \\
		$[d_{\textrm{ub}},D_{\textrm{ub}]}$ & [0.6m, 60m] &  $[e_{\textrm{con}}^{\textrm{max}}, e_{\textrm{lag}}^{\textrm{max}}]$  & {[0.5m $\quad$ 0.5m]} \\
		\bottomrule
	\end{tabular}%
	\label{tab:algoparas}%
\end{table}%

\newcommand{\ColoredBoxBrown}{\textcolor{brown}{\rule{0.5em}{0.5em}}}
\begin{figure*}[htbp]
	\centering
	\subfloat[BO training process ]
	{
		\includegraphics[scale=0.118]{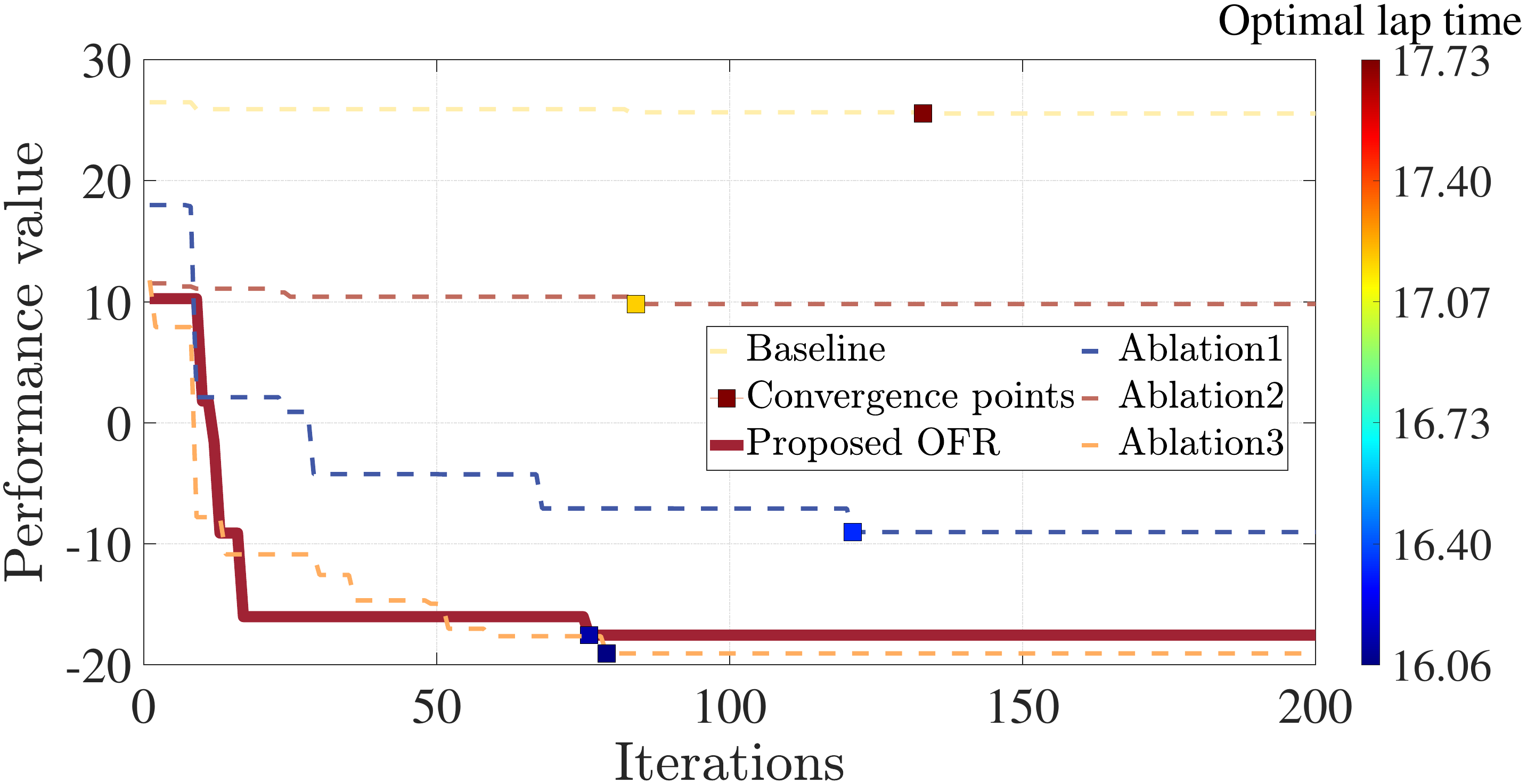}
		\label{fig:J_train}
	}
	\hspace{0.01cm} %
	\subfloat[Slip angles for racing trajectories ]
	{
		\includegraphics[scale=0.123]{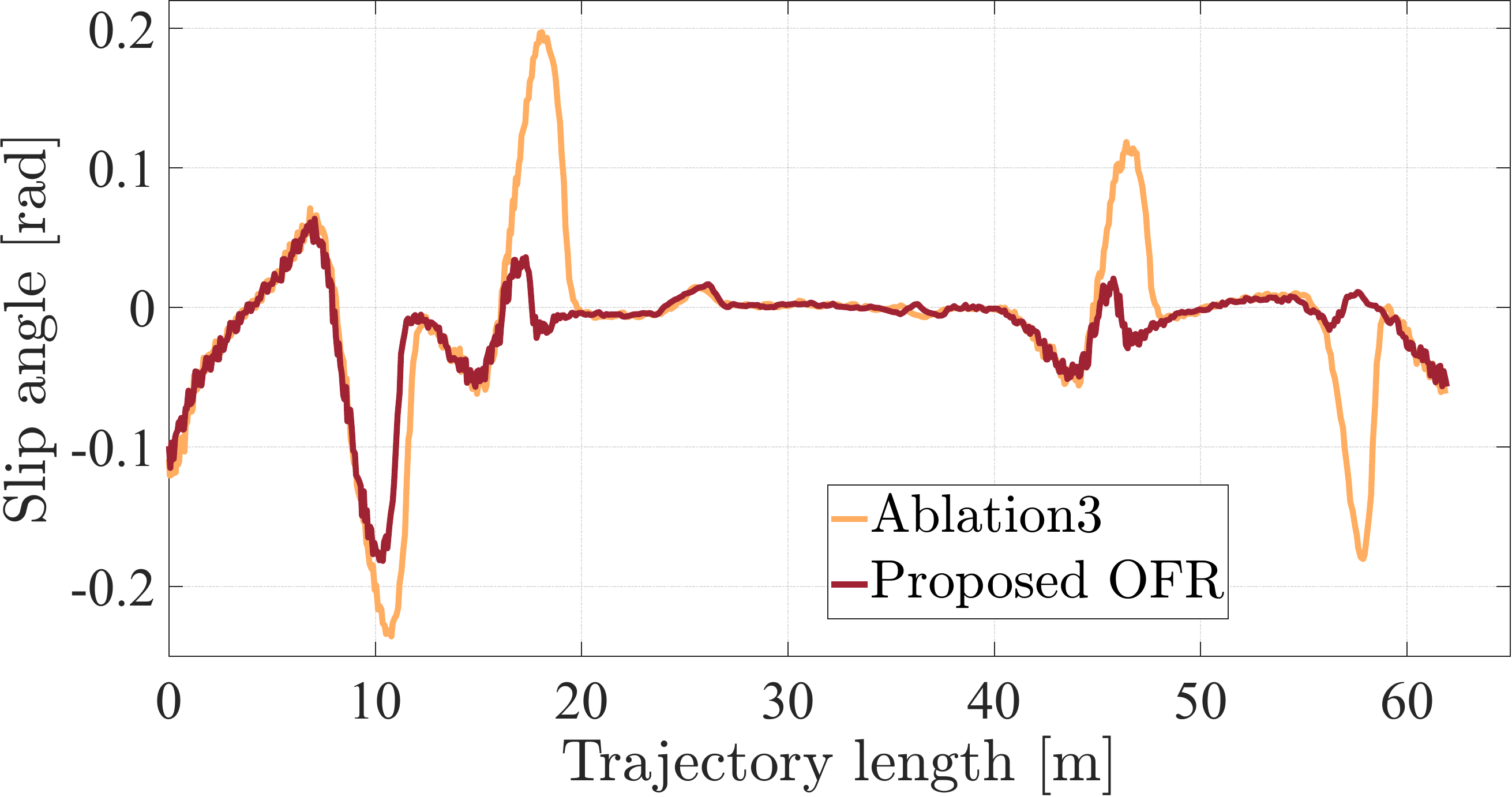}
		\label{fig:J_slipangle}
	}
	\hspace{0.01cm} %
	\subfloat[Comparison of computation efficiency ]
	{
		\includegraphics[scale=0.123]{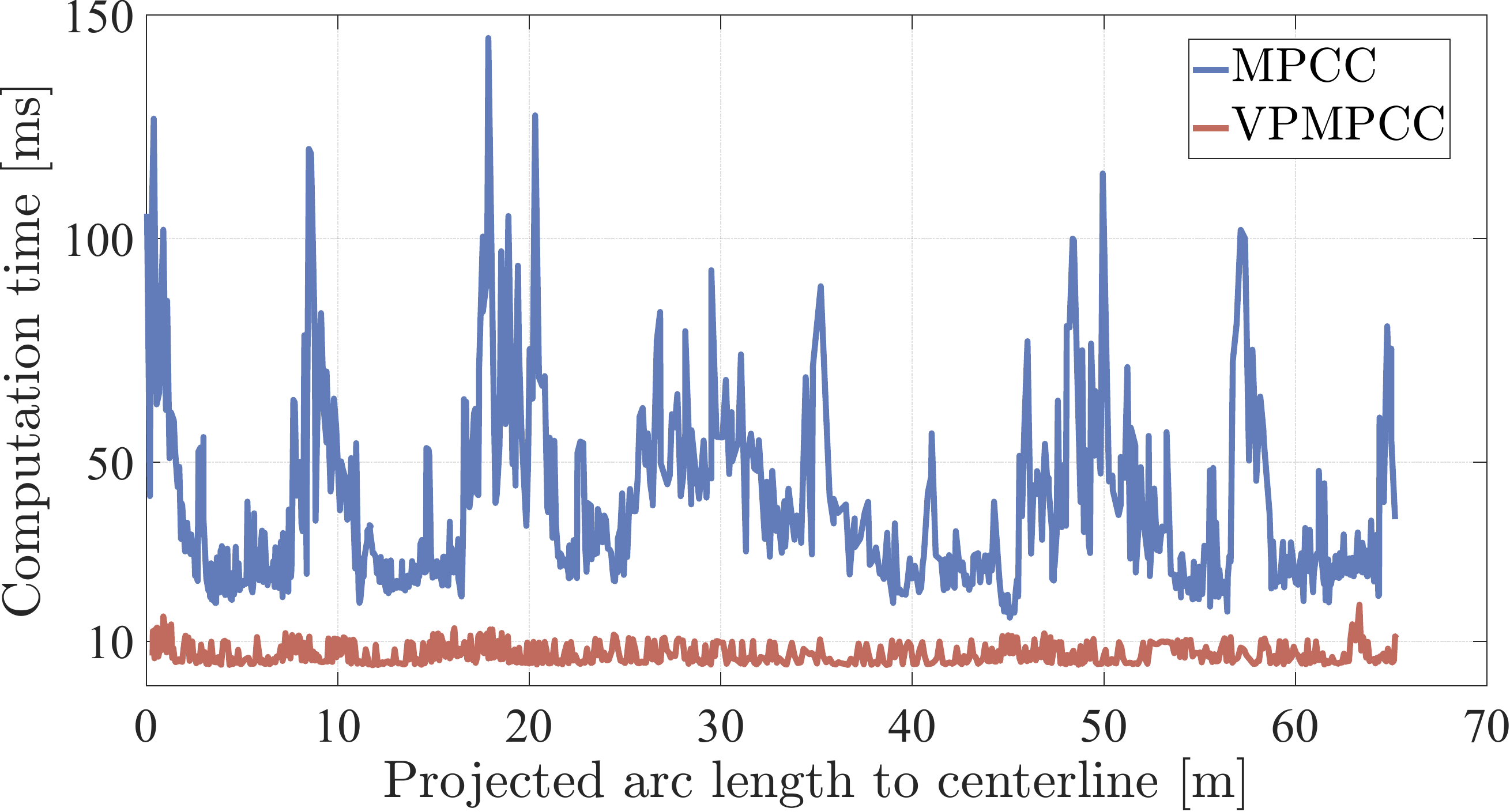}
		\label{fig:CT}
	}
	\caption{The different colored boxes \ColoredBoxBrown\; represent the optimal lap time obtained from training in Fig.~\ref{fig:J_train}. The curves indicate the current best points during training. The proposed Objective Function adapted to Racing enables BO to converge rapidly, and its optimal lap time outperforms the baseline. Ablation experiments demonstrate the effectiveness of different components of OFR for improving the training efficiency and safety of optimal trajectories. }
	\label{fig:BO}
\end{figure*}

In this section, the superiority of the proposed VPBO-RF is validated through simulations and real-world experiments. The focus is on BO training efficiency in simulations and the vehicle's mean projected velocity (Mean($v_{\textrm{p}}$)), lap time (L.T), and computation time (C.T) during real-world racing using a self-built F1TENTH vehicle (see Fig.~\ref{fig:car_comp}). A complex racetrack with four sharp corners (as shown in Fig.~\ref{fig:racetrack}) is built for simulation and real-world racing. The Cartographer ROS\cite{hess2016real} is used for mapping and localization (see Fig.~\ref{fig:map}).

\subsection{Effectiveness of the proposed OFR {in simulation}}

The runtime environment for the simulation and the F1TENTH vehicle is ROS Melodic on Ubuntu 18.04. BO training is implemented using the `bayesopt' in Matlab. The CasADi\cite{Andersson2019} serves as the solver for the optimization problem of VPMPCC. The algorithm parameters related to VPBO-RF are shown in Table~\ref{tab:algoparas}. The comparison and ablation experiments constructed in the simulation are shown below:
\begin{enumerate}
	\item Baseline (BL)\cite{frohlichContextualTuningModel2022}: The objective function is $J_{\bm{\theta}} = { T_{\textrm{lap}} + \alpha \cdot  \textrm{mean}(\mathbf{d}) } $. Because the values of $\mathbf{d}$ are small, the weight $\alpha$ is used as a trade-off with the lap time;
	
	\item Proposed OFR: The objective function is $J_{\textrm{qualified}, \bm{\theta}} = L(\mathbf{z}) + I(\mathbf{z}) + B(\mathbf{z})$;
	
	\item Ablation1: The \textbf{trajectory length} is removed from the objective function: $J_{\textrm{qualified}, \bm{\theta}} =  {L(\mathbf{z})}  
	+{B(\mathbf{z})}$;
	
	\item Ablation2: The \textbf{lap time reduction} is removed from the objective function: $J_{\textrm{qualified}, \bm{\theta}} = T_{\textrm{lap}} + {B(\mathbf{z})} +  {I(\mathbf{z})}$;
	
	\item Ablation3: The \textbf{orthogonal distance} is removed from the objective function: $J_{\textrm{qualified}, \bm{\theta}} =  {L(\mathbf{z})}  + {I(\mathbf{z})}$.
\end{enumerate}

\begin{table}[!t]
	\centering
	\caption{Statistics for verifying that the proposed OBF can improve the BO training efficiency and optimal lap time}
	\begin{tabular}{p{3.835em}cccc}
		\toprule
		Type  & \multicolumn{1}{p{4.9em}}{Optimal lap time$\,\downarrow$} & \multicolumn{1}{p{4.5em}}{Comparison} & \multicolumn{1}{p{4.9em}}{Convergence iteration$\,\downarrow$} & \multicolumn{1}{p{4.765em}}{Comparison} \\
		\midrule
		\midrule
		BL\cite{frohlichContextualTuningModel2022} & 17.73  & {-} & 133   & {-} \\
		Proposed & \cellcolor[rgb]{ .949,  .949,  .949}16.12  & \cellcolor[rgb]{ .949,  .949,  .949}-9.07\% & \cellcolor[rgb]{ .949,  .949,  .949}\textbf{76} & \cellcolor[rgb]{ .949,  .949,  .949}\textbf{42.86\%} \\
		Ablation1 & 16.33  & -7.90\% & 121 & 9.02\%\\
		Ablation2 & 17.18  & -3.09\% & 84    & 36.84\% \\
		Ablation3 & \textbf{ 16.06 } & \textbf{-9.40\%} & 79    & 40.60\% \\
		\bottomrule
	\end{tabular}%
	\label{tab:compare}%
\end{table}%

The training efficiency is calculated as $\frac{N_{\textrm{BL}}-N_{\textrm{opt}}}{N_{\textrm{BL}}}$, where $N_{\textrm{opt}}$ is iterations required for BO convergence. The percentage of racing performance improvement is calculated as $\frac{T_{\textrm{lap,BL}}-T_{\textrm{lap,opt}}}{T_{\textrm{lap,BL}}}$, where $T_{\textrm{lap,opt}}$ is the optimal lap time. The lower bound of $\bm{\theta}$ is $[ 0.1,   1,  1,  1,   1,    0.1,   1,       1 ,    0.01]$, and the upper bound is $[1,  50, 10 , 10, 10 ,  20 ,  50   ,   20  ,   0.4]$. {The final $N_p$ used to construct the MPC optimization problem is $\left\lfloor{\bm{\theta}_1 \cdot D_{\textrm{ref}}}\right\rfloor$.} The $D_{\textrm{ref}}$ is 62.8m and ${J}_{\textrm{inf}}$ is 17.6. {For Ablation1, ${J}_{\textrm{inf}}$ is 35 because the length of racing trajectories are typically shorter than that of the reference line.} The results of BO learning based on different objective functions are shown in Fig.~\ref{fig:J_train} and Table~\ref{tab:compare}.  Compared with baseline, OFR significantly improves the training efficiency of BO by \textbf{42.86\%}. In addition, the OFR converged lap time is also superior to baseline with a reduction of \textbf{9.07\%}.  Ablation1 and Ablation2 show lower training efficiency and less reduction in optimal lap time than the proposed OFR.  This demonstrates the effectiveness of the trajectory length minimization and lap time reduction components of the proposed OFR. The training efficiency of Ablation3 is comparable to that of the proposed OFR, and it achieves a shorter lap time. However, the slip angle of the optimal trajectory from Ablation3 is significantly larger than that obtained through OFR-based training (as shown in Fig.~\ref{fig:J_slipangle}). This suggests that limiting the maximum orthogonal distance of the racing trajectory can reduce slip angle. This is because OFR penalizes BO for exploring racing trajectories that deviate significantly from the reference line during training. 

\begin{table*}[htbp]
	\centering
	\caption{The statistics of continuous autonomous racing trajectories validate that VPMPCC pushes vehicle performance to its limits while maintaining stability on a challenging racetrack with four sharp corners.}
	\begin{tabular}{p{8.465em}cccccccccc}
		\toprule
		Methods$^1$ & \multicolumn{1}{p{3.665em}}{Mean(L.T)} & {Compare$^2$$\downarrow$} & \multicolumn{1}{p{3.135em}}{Mean($v_{\textrm{p}}$)} &  \multicolumn{1}{p{3.735em}}{Compare$^3$$\uparrow$} &\multicolumn{1}{p{3.135em}}{{Max($v$)}} & {Laps} & \multicolumn{1}{p{3.135em}}{Std(C.T)} & \multicolumn{1}{p{3.135em}}{Mean(C.T)} & \multicolumn{1}{p{3.135em}}{Std(Cost)} & \multicolumn{1}{p{3.135em}}{$\;\;\;$Std($a_x$)} \\
		\midrule
		\midrule
		vehicle limits\cite{heilmeier2020minimum}  & 15.88s  & {-} & 3.95 m/s  & {-} & 9.16m/s & {-} & {-} & {-} & {-} & {-} \\
		MPCC\cite{kabzanAMZDriverlessFull2020} & 30.44s  & 47.82\% & 2.16 m/s  & 54.51\% & 4.91m/s & 7     & 19.52  & 36.37ms  & 6.16  & 0.65m/s$^2$    \\
		VTAPP\cite{wengAggressiveCorneringFramework2024} & 19.90s  & 20.18\% & 3.16 m/s  & 79.82\% & 6.25m/s & 25    & {-} & {-} & {-} & 1.03m/s$^2$    \\
		VPMPCC-proposed & \cellcolor[rgb]{ .949,  .949,  .949}\textbf{ 17.05s } & \cellcolor[rgb]{ .949,  .949,  .949}\textbf{ 6.82\%} & \cellcolor[rgb]{ .949,  .949,  .949}\textbf{3.68 m/s} & \cellcolor[rgb]{ .949,  .949,  .949}\textbf{ \textcolor{red}{93.18\%}} & \cellcolor[rgb]{ .949,  .949,  .949}\textbf{6.39m/s }& \cellcolor[rgb]{ .949,  .949,  .949}\textbf{ 25 } & \cellcolor[rgb]{ .949,  .949,  .949}\textbf{ 4.64 } & \cellcolor[rgb]{ .949,  .949,  .949}\textbf{ \textcolor{red}{7.04ms} } & \cellcolor[rgb]{ .949,  .949,  .949}\textbf{ 1.49 } & \cellcolor[rgb]{ .949,  .949,  .949}\textbf{ 1.39m/s$^2$ }  \\
		\bottomrule
	\end{tabular}%
	\label{tab:statistics}%
	\begin{tablenotes}
		\footnotesize
		\item 1 The offline racing trajectory with globally minimal curvature\cite{heilmeier2020minimum}, based on ideal vehicle handling capabilities, is regarded as vehicle limits.
		\item 2 Calculated as {\tiny$\frac{T_{\textrm{lap}} \text{–} T_{\textrm{limits}}}{T_{\textrm{lap}}}$}. This metric indicates how close the lap time is to reaching the limits. The performance of MPCC is limited by corner `4'. 
		\item 3 Calculated as {\tiny$\frac{{\textrm{Mean(}v_{\textrm{p},\textrm{lap}}}\textrm{)}}{{\textrm{Mean(}v_{\textrm{p},\textrm{limits}}\textrm{)}}}$}. VPMPCC pushes vehicles to limits through the trade-off between velocity prediction and  projected velocity maximization.  
	\end{tablenotes}
\end{table*}%

\begin{figure*}[!t]
	\centering 
	\subfloat[Racing trajectories corresponding to three different methods.]
	{\includegraphics[scale=0.25]{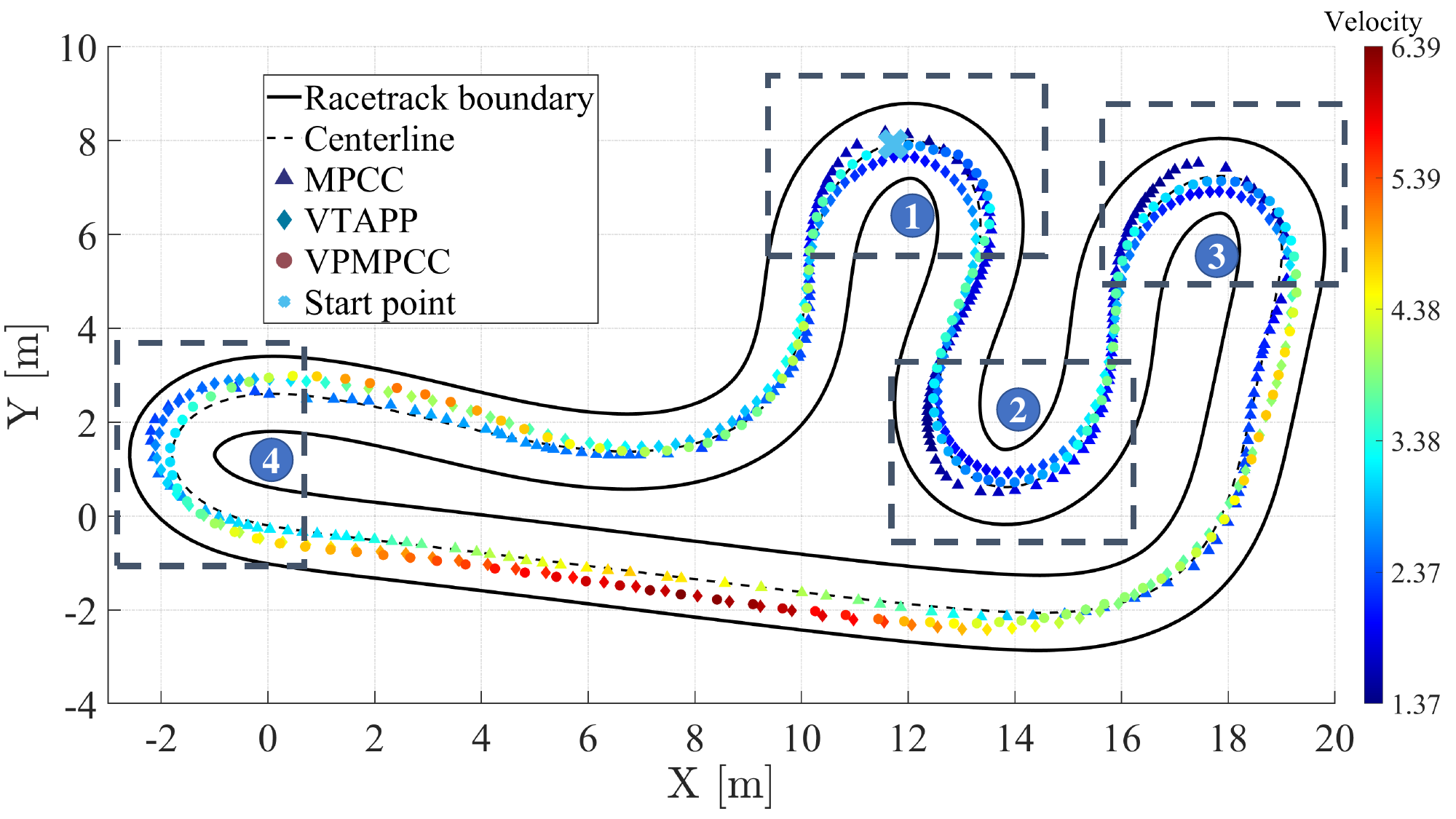}
		\label{fig:real_racingtraj}
	} 
	\hspace{0.01cm} %
	\subfloat[Longitudinal velocities corresponding to racing trajectories.]
	{
		\includegraphics[scale=0.25]{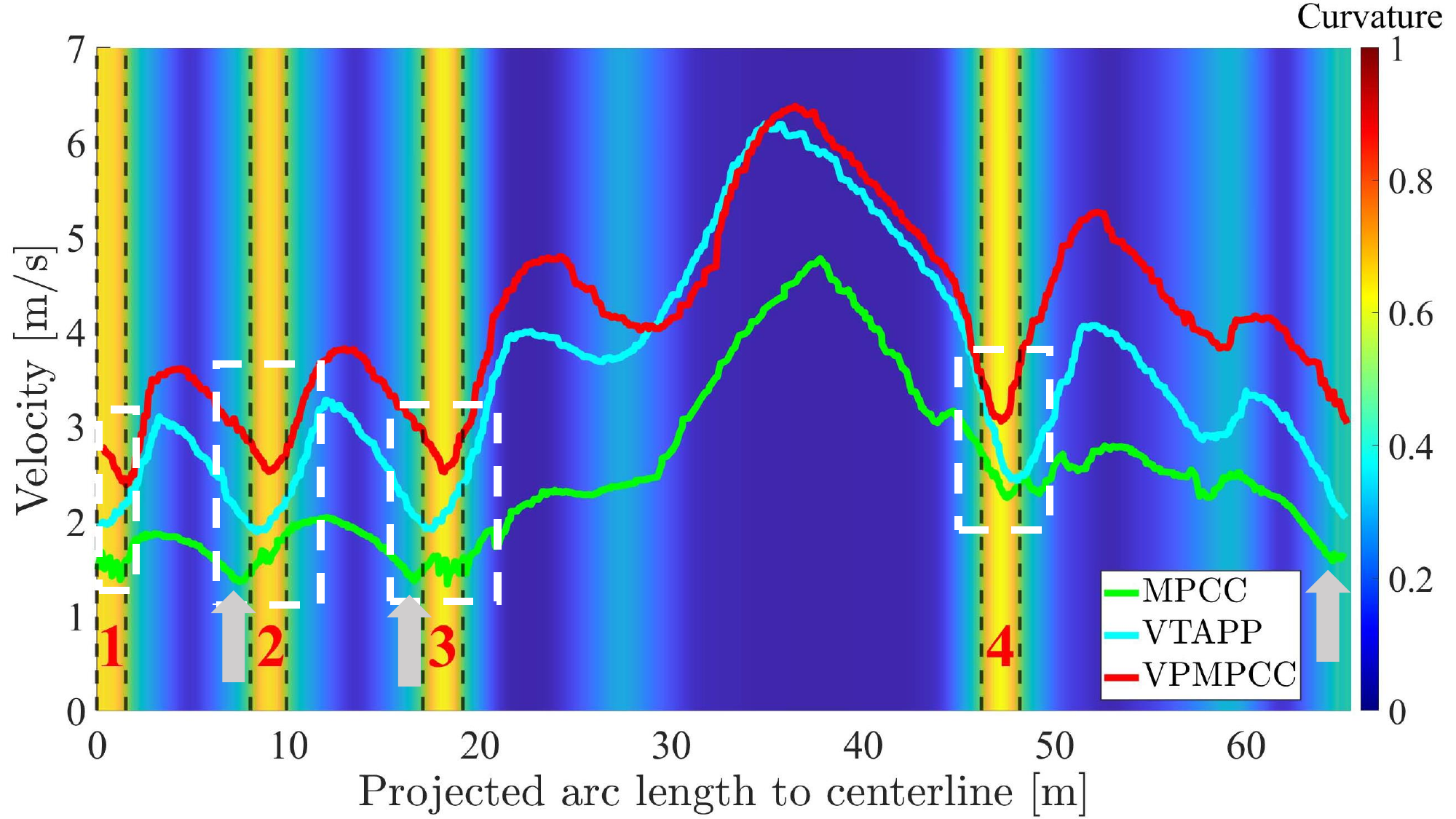}
		\label{fig:vx}
	}
	\caption{A schematic of trajectories and vehicle velocities during a single racing lap. The racing trajectory illustrates  that VPMPCC is closest to the centerline at corner `4', thereby ensuring maximum safety while maintaining racing performance. The velocity diagram shows VPMPCC achieves the highest cornering velocities, benefiting from the balance between RVP-based prediction and velocity maximization in optimal local trajectory planning.}
	\label{fig:traj_vx}
\end{figure*}

\subsection{Effectiveness of the proposed VPMPCC on {scaled vehicle}}

The MPCC proposed in~\cite{kabzanAMZDriverlessFull2020} and the Velocity-Tracking Adaptive Pure Pursuit (VTAPP) controller proposed in~\cite{wengAggressiveCorneringFramework2024} are used for comparison. The optimal parameters of MPCC are also trained through OFR-based BO, which is $[13,0,39.0,5.8,0.6,2.8,0.2,0.5,0.02]$. {The $v_{\Delta,\textrm{max}}$ is 10m/s.} And MPCC uses the racetrack centerline as the reference line, as discussed in \cite{kabzanAMZDriverlessFull2020}. The optimal parameters of VPMPCC trained by BO based on the proposed OFR is $\bm{\theta}^* = [6,3,6,3.9,1,19.0,28,15.7,0.3]$.  These three methods are used to perform consecutive laps of racing to verify the superiority of the VPMPCC method in terms of high performance and stability. The statistical data are shown in Table~\ref{tab:statistics}. The Mean($v_{\textrm{p}}$) is calculated as Mean($\frac{D_{\textrm{ref}}}{\textrm{L.T}}$).

\begin{figure}[!t]
	\centering 
	\includegraphics[scale=0.23]{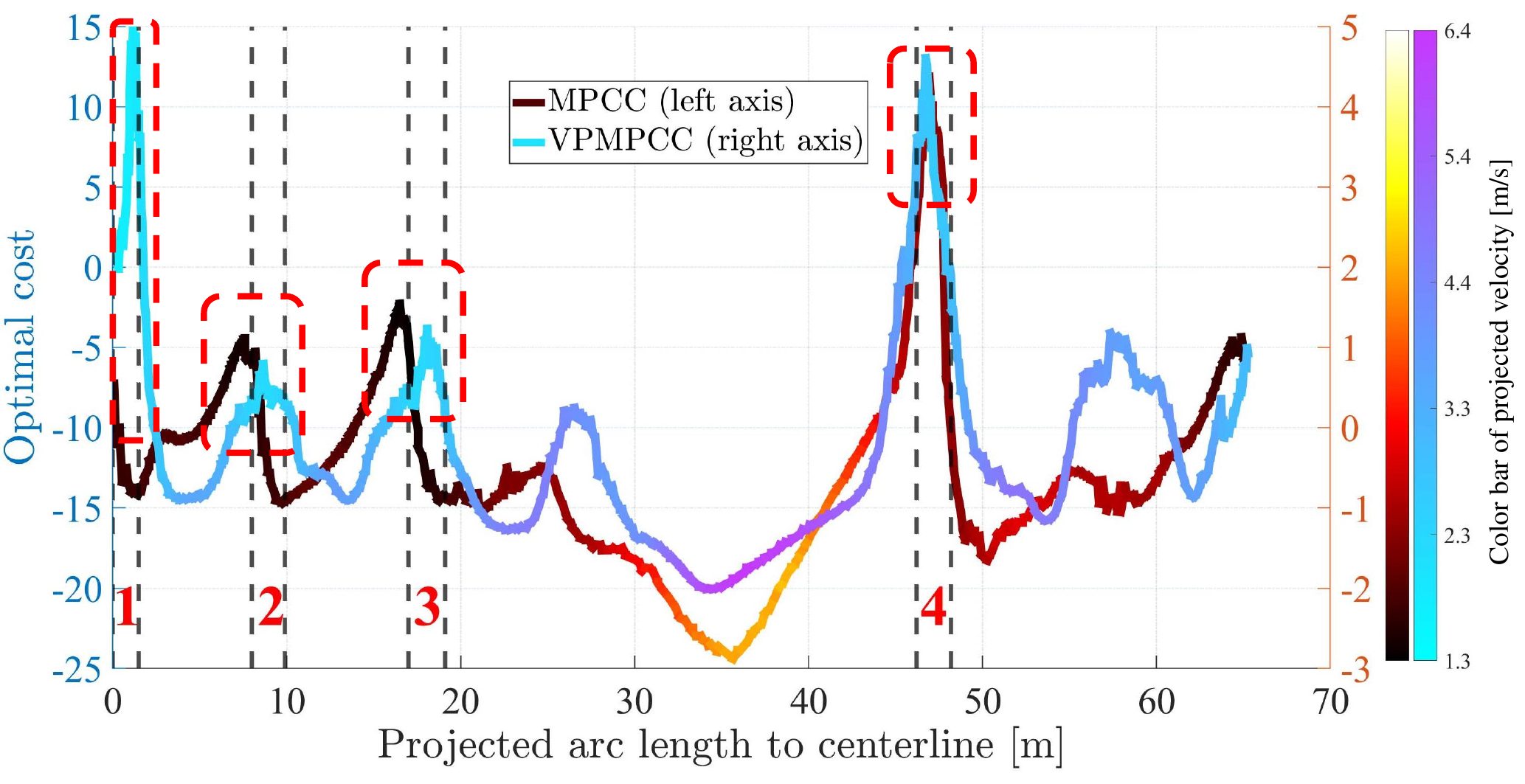}
	\caption{Demonstration of the cost for MPCC and VPMPCC in real vehicle racing. High costs indicate lower velocities. MPCC cannot plan optimal velocity profiles at corners, leading to premature deceleration at corners `1' to `3'. In contrast, VPMPCC utilizes velocity prediction to balance projected velocity maximization, maintaining optimal velocities at each corner. }
	\label{fig:cost}
\end{figure}

The mean projected velocity of VPMPCC reaches \textbf{93.18\%} of vehicle handling limits, much higher than the other methods. It also has a low computation time of \textbf{7.04ms}, which indicates that VPMPCC has high real-time performance. VPMPCC has consistently raced for \textbf{25} laps at a maximum velocity of \textbf{6.39 m/s}, demonstrating its stability. This successfully proves that VPMPCC pushes the vehicle performance to its limits with stability. Compared to VTAPP, VPMPCC maintains the effect of maximum projected velocity, thus planning local trajectories with superior velocities, especially in corners (as shown in Fig.~\ref{fig:real_racingtraj}).  As shown in Fig.~\ref{fig:vx}, during a single lap, the racing velocities of VPMPCC are significantly higher than those of MPCC. This is because MPCC employs an aggressive racing strategy, but the tracking error $\mathbf{e}_s$ is not sufficient to enable it to decelerate in time in corner `4', resulting in a possible collision. As a result, the high-velocity racing parameters of MPCC are restricted in BO training, leaving only parameters with limited racing effects. However, VPMPCC can successfully plan optimal local trajectories in corners by incorporating velocity prediction. This reduces cornering time while ensuring collision-free maneuvers, allowing VPMPCC to achieve higher global racing velocities and ultimately reduce lap time. This is also proved by optimal costs of MPCC and VPMPCC  in Fig.~\ref{fig:cost}, while MPCC successfully passes corner `4', its parameters cause a premature deceleration in corners `1' to `3' (pointed out by \textcolor{gray}{\bm{$\uparrow$}} in Fig.~\ref{fig:vx}), leading to a loss in racing performance. In contrast, VPMPCC achieves an optimal velocity profile at each sharp corner. Additionally, the cost changes of VPMPCC are smoother compared to MPCC, which further enhances the efficiency of solving the optimization problem (as shown in Fig.~\ref{fig:CT} ). 

\newcommand{\ColoredBoxGray}{\textcolor{gray}{\rule{0.7em}{0.5em}}}

\section{CONCLUSIONS}\label{sec:Conclusion}

This paper proposes a novel data-driven framework for aggressive autonomous racing that pushes the vehicle to its handling limits. By integrating velocity prediction into local trajectory planning, the vehicle's mean projected velocity reaches \textbf{93.18\%} of its handling limits. The training efficiency of BO is improved by \textbf{42.86\%} through the proposed objective function adapted to racing. By filtering out unqualified racing trajectories during BO training, the parameters trained in simulation can be directly transferred to a real-world vehicle without retraining. Future work will explore applying VPMPCC to complex multi-vehicle racing scenarios and integrating deep reinforcement learning for decision-making.





\newpage

\bibliographystyle{IEEEtran}
\bibliography{ref}

\end{document}